\documentclass[parskip=full,12pt]{article}

\usepackage{amssymb}
\usepackage{amsthm}
\usepackage{amsmath}
\usepackage{float,booktabs}
\usepackage{float} 
\usepackage{graphicx}
\usepackage{natbib}
\usepackage{url}
\usepackage{xcolor}
\usepackage{soul,color,colortbl}
\usepackage{array,multirow}
\usepackage{cancel}
\usepackage[top=0.9in, bottom=1.2in, left=0.875in, right=0.875in]{geometry}
\usepackage[linesnumbered,ruled]{algorithm2e}
\usepackage{tikz}
\usetikzlibrary{er,positioning,arrows.meta,calc}
\usepackage[labelformat=simple]{subcaption}
\usepackage{bm}
\usepackage[linesnumbered,ruled]{algorithm2e}
\usepackage{caption}
\usepackage{subcaption}
\usepackage[linesnumbered,ruled]{algorithm2e}
\usetikzlibrary{shapes,arrows,shadows}
\usepackage{bm}
\usepackage{bbm}
\usepackage{authblk} 

\usepackage{setspace}

\def\spacingset#1{\renewcommand{\baselinestretch}
{#1}\small\normalsize} \spacingset{1}

\newcolumntype{P}[1]{>{\centering\arraybackslash}p{#1}}
\newcommand*{\myfont}{\fontfamily{lmss}\selectfont}
\DeclareTextFontCommand{\textpython}{\myfont}

\DeclareCaptionLabelFormat{boldnumber}{\bothIfFirst{#1}{~}\textbf{#2}}
\title{\textbf{Privacy-Enhancing Collaborative Information Sharing through Federated Learning\\-- A Case of the Insurance Industry}}

\author[1]{Panyi Dong} 
\author[1]{Zhiyu Quan} 
\author[2]{Brandon Edwards} 
\author[2]{Shih-han Wang} 
\author[3]{Runhuan Feng} 
\author[4]{Tianyang Wang} 
\author[2]{Patrick Foley} 
\author[2]{Prashant Shah} 

\affil[1]{University of Illinois Urbana-Champaign, Champaign, Illinois, USA}
\affil[2]{Intel Corporation, Santa Clara, CA, USA}
\affil[3]{Tsinghua University, Beijing, China}
\affil[4]{Colorado State University, Fort Collins, Colorado, USA}

\begin{document}
\date{}
\maketitle

\noindent \begin{abstract}

The report demonstrates the benefits (in terms of improved claims loss modeling)  of harnessing the value of Federated Learning (FL) to learn a single model across multiple insurance industry datasets without requiring the datasets themselves to be shared from one company to another. The application of FL addresses two of the most pressing concerns: limited data volume and data variety, which are caused by privacy concerns, the rarity of claim events, the lack of informative rating factors, etc.. During each round of FL, collaborators compute improvements on the model using their local private data, and these insights are combined to update a global model. Such aggregation of insights allows for an increase to the effectiveness in forecasting claims losses compared to models individually trained at each collaborator. Critically, this approach enables machine learning collaboration without the need for raw data to leave the compute infrastructure of each respective data owner. Additionally, the open-source framework, OpenFL, that is used in our experiments is designed so that it can be run using confidential computing as well as with additional algorithmic protections against leakage of information via the shared model updates. In such a way, FL is implemented as a privacy-enhancing collaborative learning technique that addresses the challenges posed by the sensitivity and privacy of data in traditional machine learning solutions. This paper’s application of FL can also be expanded to other areas including fraud detection, catastrophe modeling, etc., that have a similar need to incorporate data privacy into machine learning collaborations. Our framework and empirical results provide a foundation for future collaborations among insurers, regulators, academic researchers, and InsurTech experts. 

\end{abstract}

\newpage
\doublespacing

\section{Introduction}\label{sec:intro}

The exponential growth of data offers substantial potential for cross-industry data sharing, gaining industry-level insights and developing standards, enhancing organizational efficiency, as well as fostering technological innovation. As an industry heavily dependent on policyholders' and external risk-related data, insurance can also leverage Artificial Intelligence (AI) advancements to fuel next-generation innovations. Prior research indicates that machine learning (ML) models can considerably improve underwriting, claim loss modeling, reserving, and fraud detection. (c.f. \citealp{baudry2019machine, blier2020machine, hanafy2021machine}). However, the insurance industry is grappling with the withholding of proprietary data, which hinders the free flow of data and collaborations between companies. Conversely, traditional ML tasks benefit from a centralized computing center or server to collect all data and the computing center is in charge of the entire life-cycle of ML tasks, ranging from data pre-processing to model deployment. However, the sensitivity and privacy of the personal data that each insurer holds make centralized data collection and model training impossible in real-life industrial applications. The lack of access to data across business divisions within a corporation and the boundaries of insurance firms make it challenging to develop comparative analysis and uncover business insights that can only be learned from data aggregation. Evidently, the lack of mature technology that enables open collaborations while safeguarding against the risks of giving away proprietary business information and client privacy data is a roadblock to next-generation innovations.

The withholding of proprietary data is rooted in ongoing discussions about the use of ML that require further investigation, despite its widespread use in various industries including insurance. Potential risks regarding leakage of customer privacy and legislation related to the evolution of ML have captured the attention of academic researchers, industry practitioners, and regulatory institutions. Some of the most pressing issues are the ownership of ML models, fairness of ML algorithms, legislation supervision, and customer/company data privacy. The transmission of ML models and model training procedures challenge the conventional definition of copyright, where it is difficult to distinguish between data providers and model trainers as the owner of ML models. Regulatory institutions need to shift their focus from solely evaluating AI industry performance gains to creating a transparent, responsible, and explainable environment. 

In this work, we propose a framework of federated learning as a privacy-enhancing solution to enable potential collaborations among the insurance industry and take advantage of state-of-the-art ML collaboration. Federated Learning, first proposed by \citet{pmlr-v54-mcmahan17a}, is a distributed approach to train an ML model that avoids the limitations and risks of centralized data storage, which is a critical prerequisite for standard ML solutions. From a business operations perspective, as summarized in \citet{LI2020106854}, federated learning is a privacy-enhancing collaborative learning technique that addresses the challenges posed by the sensitivity and privacy of data in traditional ML solutions. As a result, federated learning enables ML innovations even when data centers or computing servers cannot collect datasets physically. With the rise of federated learning, researchers and practitioners have joined forces in various multidisciplinary federated learning projects to achieve goals that were once believed to be impossible. See \citet{niknam2020federated}, \citet{rieke2020future}, and \citet{Pati2022}. Among all potential collaborations, our work focuses on the application of federated learning in insurance claim loss modeling, addressing two of the most pressing concerns: data volume and data variety.

We propose that federated learning would be the next evolution in the insurance industry, solving data challenges with unprecedented effectiveness and efficiency by sharing knowledge and expertise while enhancing privacy. For instance, through the partnerships between different insurance companies, and among insurance companies and InsurTech companies, each private ``data owner" trains a model locally using only their own data, the parameters of which are then shared with the central ``aggregation server" to create a consensus model with accumulated knowledge from all ``data owners" to create a more complete picture of risk assessment of policyholders and industry-level of insights. Such multi-institutional collaboration without sharing data among the collaborators is of great value to insurance companies, industrial associations, governments, and regulators.

\section{Data Privacy-Enhancing in Insurance Industry}\label{sec:use-case}

The insurance industry, which mitigates and manages societal risks, exerts a notable influence on society and the global economy. Insurance companies provide financial protection to individuals, businesses, and communities against various types of risks, such as death, injury, illness, property damage, and liability, by collecting premiums from policyholders and using the funds to pay out claims in the event of a direct or indirect financial loss. The insurance industry is characterized by its complex business structure, which encompasses various lines of business such as life, health, auto, property, liability, and commercial insurance. Furthermore, the life cycle of insurance products, including underwriting, pricing, claim reporting, claim processing, and fraud detection, adds to the complexity of the industry's structure. The insurance industry has evolved over time and gradually adapted to the modernization of society, with the development of innovative insurance products, advanced risk management techniques, and sophisticated IT systems. In recent years, the industry has undergone significant changes due to technological advancements, changing customer demands, and regulatory changes, making the insurance industry a dynamic and challenging field that requires continuous innovation and adaptation.

The insurance industry is highly regulated by country and, in some cases, by state or province, which aims to protect policyholders. Furthermore, insurance companies, as data owners, do not necessarily have all the competencies or budgets to develop various proof of concept (POC) projects that use AI technology. Legal constraints and technical barriers prevent data owners from further improving the ML models that can be trained on their limited data. To resolve these issues, we propose a research project to demonstrate that in a federated learning environment, a non-profit third-party university research group can provide strong techniques and act as a centralized hub to promote federation among insurance companies and InsurTech companies, and significantly improve the prediction performance of actuarial loss models. Meanwhile, data owners do not need to share their private data with other parties.

The demonstration of the practicality of federated learning requires proprietary data from multiple insurance companies and InsurTech firms, which is often challenging for researchers. Gratefully, we have secured a collection of such proprietary data from multiple insurance industry partners to showcase the framework and methodology of federated learning in the insurance industry. Previously, we established a research framework for collaboration between universities and insurance industry partners. The collaboration enriched the existing dataset for research by providing proprietary information from the industry. This collaboration is unprecedented in finance and business in general.

Based on such a collaborative relationship, we propose applying ML, specifically supervised learning, to improve the partners' in-house models in a hypothetical federated learning environment. As a result, we illustrate that real-life data created from the InsurTech innovation, which introduces data variety (additional information), can help significantly improve the underlying insurance company's in-house models. In addition, uniting multiple insurance companies through federated learning will enhance every collaborates' in-house models regarding the same line of business or the same business purpose. This innovation can be universally applied in the insurance industry in broader contexts.

\subsection{State of Claim Loss Modeling}\label{subsec:loss-modeling}

Claim loss modeling, referred by \citet{klugman2012loss}, is the term denoting the procedures of forecasting future claims reported by customers or policy costs through previous experience in the insurance industry, which is a critical indicator for insurance product pricing, underwriting, and thus crucial for insurance companies. In the case of pricing, the future claims predicted by the claim loss model represent the essential costs that companies are expected to prepare. These essential costs, denoted as \textit{net premium}, implies an infimum of the premium that the insurance companies need to charge from customers. One of the common practices that insurance companies take to estimate the \textit{gross premium} (the estimated price insurance companies should charge from policyholders with the consideration of companies' operation expenses, underwriting costs, commissions for insurance agents and other forms of costs and profit) is to evaluate the \textit{gross premium} through \textit{net premium}. Thus, claim loss modeling is one of the most important components of the insurance business workflow. In addition to pricing, the claim loss models also serve as the analytic identifier of business trends to make informed decisions and satisfy regulations for reserve and solvency purposes. The more accurate the claim loss models insurance companies hold, the more precise the future claim losses can be predicted, which, from the perspective of policyholders, allows the improvement of pricing fairness and, from the viewpoint of insurance companies, reduces the operation risk and enhances the competitiveness in the market. Insurance companies, driven by the motivation to increase profits and improve risk management, are always interested in techniques that help the capability of claim loss models.

In traditional insurance practice, the claim loss modeling can be deciphered by frequency and severity, which denotes the number of claims and cost of each claim, respectively. Following the notations from \citet{klugman2012loss}, to address the problem of claim loss modeling and estimate the aggregated loss $S$, one of the typical practices in the insurance industry is to separate the estimation of frequency $N$ and the independently and identically distributed severity, $Y_{i}$, ($i=1,2,...,N$), and project the expected aggregated claim loss following
$$
\mathbb{E}[S]=\mathbb{E}[N]\mathbb{E}[Y_{i}]
$$
Alternatively, aggregated claim loss $S$ can be measured by a compound distribution that considers both frequency and severity so it can be expressed in the form
$$
S=Y_{1}+Y_{2}+...+Y_{N}
$$
One of the traditional approaches to modeling this aggregated claim loss is using a standard mixture model which is compound Poisson-Gamma distribution, specifically Tweedie distribution \citep{jorgensen1994fitting}, where the frequency of claims follows Poisson distribution and the severity of each claim can be described by Gamma distribution. However, with advances in AI, powerful ML models can be trained to forecast aggregated claim losses directly with improved prediction performance. Research work with applications of ML techniques in the insurance sector has emerged in the last few years. See \citet{GUELMAN20123659}, \citet{baudry2019machine}, \citet{blier2020machine} and \citet{hanafy2021machine}.

\subsection{Shortage of Data}\label{subsec:shortage-data}

Data is critical in ML because it provides the foundation for creating accurate and effective models. Data is particularly important in insurance because it is used for accurate risk assessment, pricing, managing risk, and identifying fraud, among others. Structured tabular datasets are organized by rows and columns. In the following, we refer to each row as one observation, which, in the reflection of insurance claim datasets, corresponds to the information of a policy/policyholder, and each of the columns denotes either a feature (risk factor) that insurance companies collected to describe the policy and identify potential risks, or a label indicating a target value (observed claim loss) the model is attempting to predict. A structured tabular dataset, in the following, denotes $\mathcal{D}=(\textbf{X}, \textbf{y})=(\textbf{X}_{i}, y_{i})$, ($i=1, 2, ..., n$), where $\textbf{X}$ refers to features, $\textbf{y}$ refers to labels, and each of the total $n$ observations can then be correspondingly characterized by the pairs of $(\textbf{X}_{i}, y_{i})$. Each of the inputs $\textbf{X}_{i}$, can be represented as $\textbf{X}_{i}=(X_{i1}, X_{i2}, ..., X_{ip})$ where each of $X_{ij}$, ($j=1, 2, ..., p$) indicates the cell values of total $p$ features stored in the dataset. Thus, the inputs $\textbf{X}$ comprise a $n\times p$ matrix, where $n$ refers to the number of observations and $p$ denotes the number of features utilized. 

In the format of such structured tabular datasets, the data quality is significantly impacted by $n$ and $p$. The foundation of insurance businesses is built upon the capability of accurately quantifying risks (potential car damages in auto insurance, costs of a lawsuit in liability insurance, mortality in life insurance, etc.) utilizing all collected information. Two of the most promising improvements that insurance companies can benefit from through ML collaboration correspond to the increase in the number of observations ($n$) and the number of features ($p$). The claim events, usually referred to as ``accidents", are inherently infrequent, but have a significant impact on how risky policyholders can be differentiated from safe customers. 

\textbf{Lack of Data Volume}: In the claim loss experience of the insurance industry, the proportion of policyholders who submit at least one claim to insurance companies can be as low as 10\%, and in a few business lines that cover only catastrophic crises or extreme events, the rate can be even lower than 0.1\%. As a result of such rarity, the risk characteristics that help identify the risks can be under-represented by the occurred claim events or are not producing noticeable signals to be captured by claim loss models. Thus, without any collaboration, most insurance companies are always in a short supply of claim events to allow their claim loss models to better learn from the behaviors of risky policyholders. This is particularly true for insurers entering a new market without claim experience. Furthermore, one of the common techniques for solving a similar problem of imbalance is sampling, whose primary objective is to generate synthetically and statistically similar claim events or eliminate part of no-claim policyholders from the database to reshape the dataset into a balanced one. However, the sampling techniques are generally considered an unacceptable approach in the insurance domain as they distort the distribution of policyholders' behaviors. Synthetic observations, though maybe statistically realistic, may not be a true representation of policyholders. Due to the limitation of sampling techniques in the insurance industry, a proper ML collaboration solution is needed to solve the issue. 

In the case of collaboration among multiple insurance companies, we assume that, due to the homogeneity of collaborators as insurance companies, their data share a common set of features, as well as a common label. The observations at all companies in this scenario could hypothetically be combined to form a global dataset in which the individual company datasets are horizontal partitions because these partitions are defined by subsets of rows (observations). A technical solution is desired that can simulate the act of training on the global dataset without actually requiring the companies to expose their data.

As collaboration in this scenario increases the number of observations, we argue that this collaboration increases the data volume. Generally, the more companies participating in this collaboration, the larger the data volume, and correspondingly, it can provide a better characterization of risks from an industry-level perspective and avoid biases in the model that can lead to inaccurate predictions or recommendations. Such a collaboration is referred to as Horizontal Federated Learning (HFL).

\textbf{Lack of Data Variety}: Another noteworthy potential improvement coming from the collaboration of the insurance industry lies in the expansion of the feature space. To accurately predict future claim losses, as many risk factors as possible should be considered. For instance, in addition to the risk factors traditionally used for modeling small business claim losses, including but not limited to business categorization, annual revenue, and employee size, more risk factors can be used through the evolution of new technologies such as the Internet of Things (IoT) \citep{spender2019wearables}, telematics \citep{handel2014insurance} and social media \citep{mosley2012social} of the businesses in question. The true risk factors may lie in those additional features that traditional insurance companies do not consider in the modeling process. Furthermore, the cross-industry collaboration between insurance and related fields like banking and InsurTech data vendors can be beneficial, as the cross-section of small business policyholders among these industries creates an opportunity to characterize policyholders better than traditional underwriting procedures. 

For two companies where the scope of the collaboration is to improve the capability of insurance claim loss modeling, one is an insurance company that is in possession of a dataset with features and labels corresponding to observations on a set of policyholders, whereas the other one, an InsurTech company, holds a dataset of additional (new-technology-empowered) features collected on the same set of policyholders as those known by the insurance company. Note that the InsurTech company has no label information for this scenario. 

Due to the common set of policyholders associated with the two company datasets, the features of the InsurTech company, and features and labels from the insurance company could hypothetically be combined to form a `global' dataset of which the individual company datasets are `vertical' partitions due to the fact that these partitions are defined by subsets of columns (features, label). Again, a technical solution that can simulate the act of training on this global dataset without actually requiring the companies to combine their data is desired. In this scenario, the collaboration between an insurance company and an InsureTech company increases data variety since the additional features result in improved claim loss modeling. Such a collaboration is referred to as Vertical Federated Learning (VFL). 

\subsection{Privacy Concerns}\label{subsec:privacy-concern}

Despite the fact that the insurance industry has long been concerned with the shortage of both the volume and the variety of data, wide-scale solutions have not been deployed that can effectively eliminate privacy concerns of sharing data to a central location. Though performance enhancements would surely come from collaborative data sharing to centralize such ``data silos" \citep{LI2020106854}, there is an inherent reluctance to do so. Insurance companies, which are central to the practice of risk forecasting and management, adopt a cautious and conservative approach in embracing ML collaboration due to the protection of customer privacy and the safeguarding of embedded confidential business strategies. Thus, techniques enabling privacy-enhancing ML collaborations that can simulate the advantages of data sharing, while providing protections against data exposure, can benefit the insurance industry by offering a resolution to the persistent shortage of data volume and variety.

In this work, we utilize the technique of FL towards privacy-enhancing ML collaboration. Insurance companies can take advantage of ML collaboration while mitigating the exposure of their raw data. FL is a technique for training an ML model across disjoint data owners, while allowing all collaborators to keep their raw data within their own computing infrastructure rather than sharing data with other collaborators. Only model updates and evaluation metrics are shared. Though techniques have been demonstrated to show that the model updates shared during FL have the potential to leak information about the raw data itself \citep{zhu2019deep}, the FL framework utilized in this study addresses such threats. We use the OpenFL\footnote{Full reference of OpenFL. Retrieved from: https://github.com/securefederatedai/openfl} framework to run the experiments in this study. OpenFL (developed by \citet{openfl_citation}) is designed to mitigate information leakage during FL through a combination of privacy-enhancing algorithms (e.g., final model leakage mitigation) and the use of trusted execution environments to control access to the model updates and evaluation metrics as well as to ensure all algorithmic enhancements are performed with integrity. 

We explore the utilization of HFL to evaluate the impact of an increase in data volume through collaboration among insurance companies, and VFL to evaluate the impact of an increase in data variety through collaboration between insurance and InsurTech companies. With the added privacy protections provided by frameworks such as OpenFL, insurance companies participating in federated model training can benefit from enhanced business intelligence due to increases in data volume and/or data variety, while mitigating privacy exposure that would come from data sharing to achieve the same goal.

\section{Methodology}\label{sec:model}

\begin{figure}[h!]
  \centering
    \includegraphics[width= 0.3\linewidth]{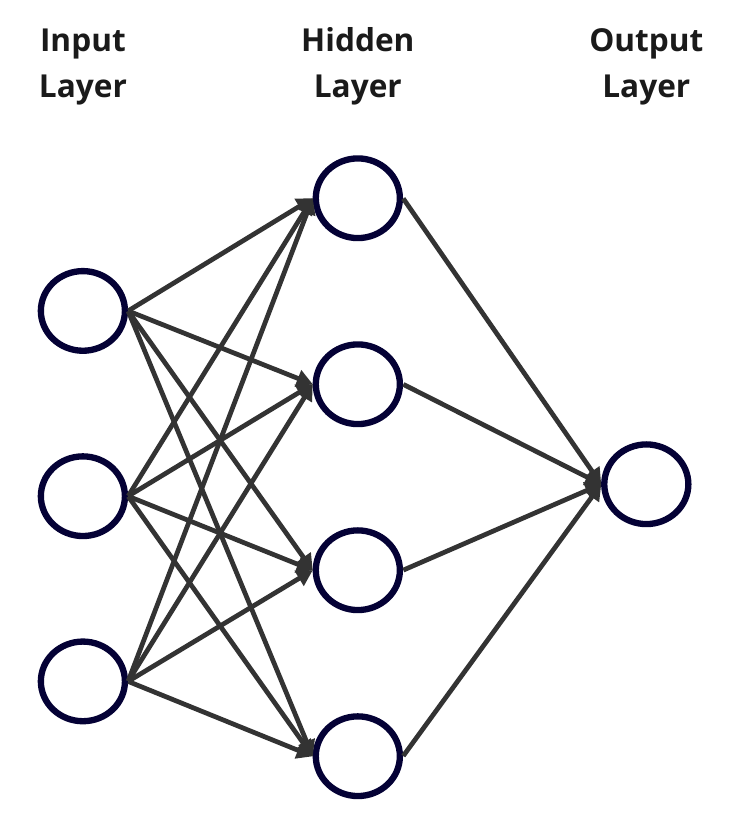}
  \caption{Neural Network}
  \label{fig:NN}
\end{figure}

The empirical experiments conducted in this work utilize the framework of HFL/VFL with the model architecture of Neural Networks. Neural Networks (NN), as summarized by \citet{muller1995neural}, is an ML model architecture inspired by the structure and functionality of neural cells in the human central nervous system. The NNs comprise connected artificial neurons, as illustrated in Figure \ref{fig:NN}. NN has proven to improve businesses and even offer possibilities for industry advancement. The backward propagation technique employed to update NN parameters can be conveniently adjusted to become compatible with the FL framework, while the design of ML architectures like Decision Trees and Support Vector Machines demands proper modification to fit into the FL framework. The promising model performance and ease of adaptation of NN are two dominant reasons for architecture selection.

In FL, two categories of participants exist: collaborators and the central server. The collaborators are in charge of, by order of each round of training, receiving the up-to-date aggregated model parameters, local model updates exploiting private data on their computing devices, and sending locally-updated model parameters in preparation for model aggregation. The central server differs drastically from the conventional one in that it does not hold any private information from collaborators or participate in any model updates in need of those confidential data. The central server in FL only takes the role of communication to receive local model parameters from all collaborators, perform certain aggregation procedures utilizing only the model parameters and send aggregated model parameters to each of the collaborators. Since the central server is only in possession of model parameters throughout the entire training process, the privacy of raw data is preserved by local collaborators.

The common categorization of FL solutions, as suggested by \citet{Yang2019}, lies in the differences in data partition and distribution through feature and observation space. In the following work, we will focus only on the cases where collaborators share a common set of features or policyholders for observation, corresponding to HFL and VFL, respectively.

\begin{figure}[h!]
  \centering
    \includegraphics[width= 0.4\linewidth]{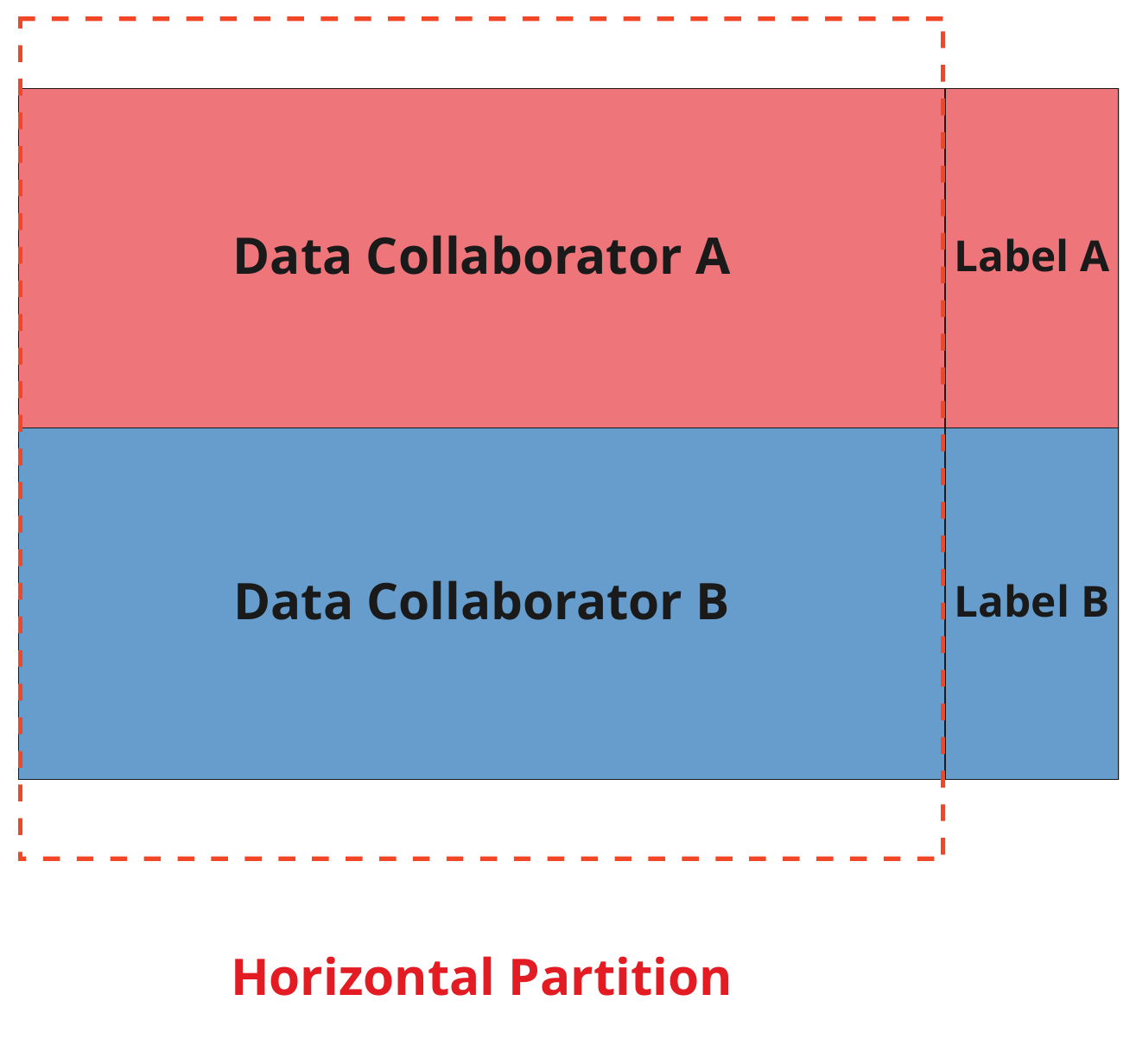}
  \caption{Horizontal Data Partition}
  \label{fig:HFL}
\end{figure}

As illustrated by two-party HFL collaboration in Figure \ref{fig:HFL}, HFL refers to the case where there exist common features between data collaborator A and B. HFL, by its design, solves the problem of heterogeneity in observation space with the condition of homogeneity in feature space. This accommodates the collaboration of participants in possession of the same data format.

\begin{figure}[h!]
  \centering
    \includegraphics[width= 0.5\linewidth]{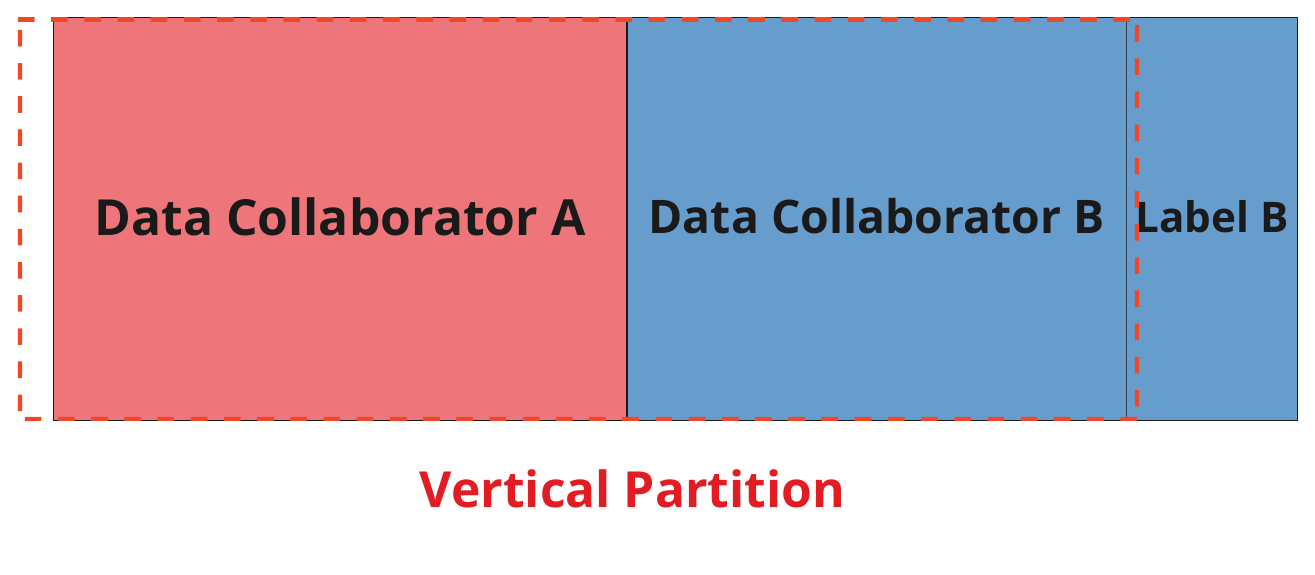}
  \caption{Vertical Data Partition}
  \label{fig:VFL}
\end{figure}

Similar to the categorization of HFL, VFL, as demonstrated in Figure \ref{fig:VFL}, denotes the vertical partition where partial observations are shared among data collaborators related to the same observation subjects. Specifically, different from HFL, where all collaborators are in possession of labels, in the scheme of VFL, usually, only one collaborator holds labels of interest while other collaborators solely contribute features unique to its field. VFL applies to the case of homogeneity of observations but heterogeneity of features. In real-life applications, VFL satisfies the demand for multi-disciplinary collaborations.

\section{Empirical Experiments}\label{sec:exp}

\subsection{Data}

Two insurance companies have shared their claim loss data in commercial lines with their rating factors. Specifically, one company shared their claim loss data for Business Owner’s Policy (BOP) insurance (which we refer to as \textit{company BOP} and whose dataset is denoted by $\mathcal{D}_{BOP}$). Most BOP bundles contain essential coverage of property insurance and liability insurance, and we extract liability parts from the BOP dataset for our purposes. A total of 392,726 observations are contained in $\mathcal{D}_{BOP}$, each row of which contains 26 feature values for a single observation (i.e., small business policyholder) together with a label indicating a claims loss value (which could be no loss at all). Likewise, the company we refer to as \textit{company GL} shared its claim loss data in General Liability (GL) insurance, whose dataset we denote by $\mathcal{D}_{GL}$. GL insurance, also called business liability insurance, is an insurance product designed to protect businesses from various claims, including bodily injury, property damage, personal injury, and others that can arise from business operations. The data shared by \textit{company GL}, which we denote $\mathcal{D}_{GL}$, consists of 210,857 observations comprised of 39 features and a claims loss label, each observation of which corresponds to a small business policyholder.

The InsurTech company, Carpe Data\footnote{\url{https://carpe.io/}}, provides business-related information from multiple data sources on the same policyholders corresponding to the insurance company data above. We categorize the additional information into five categories: Business Information, Risk Characteristics, Customer Reviews, Website Contents, and Carpe Index. We refer to this company as \textit{company CD} and denote its dataset (of features only) by $\mathcal{D}_{CD}$. There are 555 features for each observation within $\mathcal{D}_{CD}$, and there are a total of 603,583 observations, each of which provides the values for all 555 features corresponding to a small business policyholder corresponding to a row of data in either of the insurance company datasets $\mathcal{D}_{BOP}$ or $\mathcal{D}_{GL}$ described above.

The $603,583$ rows of $\mathcal{D}_{CD}$ split into 392,726 rows of observations that correspond (row by row)
to the 392,726 rows of $\mathcal{D}_{BOP}$ (providing 555 additional features describing each of the policyholders contained within), as well as 210,857 rows of observations similarly corresponding to the rows of $\mathcal{D}_{GL}$.

\subsection{Experiment Design}

In the empirical study, although there are various insurance scenarios that can potentially benefit from the application of the FL framework, we specifically focus on the impact of privacy-enhancing collaborative learning on insurance claim loss modeling in both data volume and data variety aspects. As introduced in Subsection \ref{subsec:loss-modeling}, claim loss modeling is a critical topic in all insurance business lines since it estimates future claim losses that will incur and be reported by policyholders and serves as a lower bound of the product price to guarantee the company's solvency reserve and capability of profit. As such, it has become a popular topic among insurance academic research and industrial operation, ranging from traditional statistical distribution estimations summarized in \citet{klugman2012loss}, to the applications of ML innovations suggested by \citet{GUELMAN20123659}. 

Our work employs the Feedforward Neural Network (FNN) structure introduced in \citet{SVOZIL199743} as our learning model architecture. FNN, as one of the initial NN structures, comprises a series of fully-connected linear layers and non-linear activation functions. In addition, the experiments are conducted within the open source framework of OpenFL\footnote{Full reference of OpenFL. Retrieved from: https://github.com/securefederatedai/openfl} developed by \citet{openfl_citation}, which is a popular open-source federated learning framework combining common FL algorithms with secure channeling solutions.

In our HFL experiments exploring the impact of increased data volume, we perform HFL between two collaborators who each hold independent observations but the same set of features. Here we compare the performance of each collaborator's `local' model trained only on their own data, with the performance of the HFL model resulting from federated training.

In our VFL experiments exploring the impact of data variety, we perform VFL between a collaborator that holds some features and labels on a set of observations, and another collaborator that holds additional features for each of these sets of observations. Here we compare the performance of the `local' model resulting from the first collaborator training on its data, with that of the VFL model resulting from federated training across the two data sets.

\subsection{Impact of Data Volume}\label{subsec:exp-hfl}

As introduced in Subsection \ref{subsec:shortage-data}, some of the insurance business lines suffer from the shortage of claim events due to limited occurrence, which prevents companies from comprehensively studying the behaviors of risk characteristics. However, such a shortage of claim events, through a collaboration among insurance companies sharing the same concerns and selling similar insurance products, can be solved by the technique of HFL where a model can be trained across the union of the data collected by multiple companies. Here, we define \textit{collaborator A} as holding the 392,726 labels from the rows of $\mathcal{D}_{BOP}$, together with the 555 features from $\mathcal{D}_{CD}$ corresponding to those same rows, and define \textit{collaborator B} as holding the 210,857 labels from the rows of $\mathcal{D}_{GL}$, together with the 555 features from $\mathcal{D}_{CD}$ corresponding to those rows. 

The training at each collaborator uses only the local dataset held by the collaborator. Claim loss modeling can be considered as a supervised regression task, for which we utilize the architecture of FNN with each collaborator training with the identical architecture. We utilize the \textit{FedAvg} algorithm \citep{pmlr-v54-mcmahan17a}, where the federation comprises only two insurance collaborators and the features leveraged by the two collaborators are identical. The cooperated training procedure consists of iterative local training and aggregation communications. The local training in HFL, for each of the collaborators, is equivalent to one epoch of local training without the FL framework, while the rounds of communication distinguish the FL from local training through an aggregation of the two local model updates for each round. During the local training, both collaborators, in our experiment settings, optimize the aggregated model from round $t$, represented by $F_{t}$, into $F_{t+1}^{1}$ and $F_{t+1}^{2}$ respectively, which are then averaged during aggregation to generate the global model for round $t+1$ according to
$$
F_{t+1}=\dfrac{F_{t+1}^{1}+F_{t+1}^{2}}{2}
$$
This new global model is then shared to both collaborators to use as the initial model state for the next round of training.

\subsection{Impact of Data Variety}\label{subsec:exp-vfl}

Proposed in Subsection \ref{subsec:shortage-data}, due to the complexity and rarity of claim events, finding the features that can completely characterize the risk is a difficult challenge to the insurance industry and companies are exploring all feasible options that can expand the feature space their claim loss models can train on. As one of the solutions, VFL, allows for collaborative training using distinct feature sets (each held by a different collaborator) on a common set of observations. In our empirical experiments, there are specifically two collaborators that participate in the privacy-enhancing collaboration: an insurance company and an InsurTech company. InsurTech companies can provide alternative features to those the insurance companies have in their data, which can contain additional signals used to potentially improve predictions of the claim loss model. With VFL, collaborative training can proceed while privacy issues related to raw data exposure that would be present with direct data sharing can be mitigated. Here, we define \textit{collaborator A} as holding the 26 features and label for all 392,726 rows of $\mathcal{D}_{GL}$, and \textit{collaborator B} as holding the 555 features from $\mathcal{D}_{CD}$ corresponding to those rows.

VFL is distinguished from HFL in Subsection \ref{subsec:exp-hfl} by the fact that the data sets at the two collaborators hold distinct features on common observations (policyholders), with one collaborator additionally holding the label. In this case, we do not have a shared model structure utilized at the two collaborators. Instead, we adapt the structure of SplitNN proposed by  \citet{ceballos2020splitnn} to perform training of the full architecture used in the HFL experiments. In our experiment setting, \textit{collaborator B} participates as a \textit{feature worker} while \textit{collaborator A} engages as a \textit{feature-label worker}. The training progresses with a combination of parallel and sequential operations. First, \textit{collaborator A} and \textit{collaborator B} feed a batch of their respective features (synchronized to correspond to the exact same observations) to locally-hosted models $F_{1}$, $F_{2}$ (respectively, split from the head of the larger NN trained on during HFL) to generate intermediate feature representations that are then concatenated and transmitted back to the label holder, \textit{collaborator A} in our experiment, who utilizes them to predict the claims through the final tail segment of the architecture. Upon calculation of the objective loss at the label worker, backward propagation occurs in order to update the tail model segment, and the backpropagation results are returned to both feature collaborators to further backpropagate through $F_{1}$ and $F_{2}$ and complete the update of those models for the batch. In contrast to HFL where all collaborators share the final model,  individual feature and label workers hold and update independent segments of the model.

\subsection{Empirical Results}\label{subsec:res}

In accordance with the experimental design outlined in Subsection \ref{subsec:exp-hfl} and \ref{subsec:exp-vfl}, this subsection presents an analysis of the impact of privacy-enhancing machine learning collaboration on data volume and data variety. The results from experiments indicate that, while addressing data privacy concerns, the FL framework can potentially improve the accuracy of claim loss event prediction for insurance companies. The performance of the ML model, in our experiments, is evaluated by metrics of Percentage Error (PE). This metric embeds meaningful business intuition that measures the portfolio level accuracy.

The Percentage Error (PE) can be expressed as
$$
PE(\textbf{y}, \hat{\textbf{y}})=\dfrac{\sum_{i}(y_{i}-\hat{y}_{i})}{\sum_{i}y_{i}}
$$
where $\textbf{y}$ denotes the true values and $\hat{\textbf{y}}$ refers the predicted values. Given the definition, $PE(\textbf{y}, \hat{\textbf{y}})$ has a range of $\mathbb{R}$ and smaller the value of $|PE(\textbf{y}, \hat{\textbf{y}})|$, the more accurate the predictions are. It's important that PE offers insights into forecasts at the portfolio level, indicating the capability of sustainability and solvency of insurance companies' estimated claim loss reserves and is a crucial indicator in industrial practices.

\begin{table}[!ht]
\centering
\begin{tabular}{c c c rrrr} \hline
Collaborator & Split & Mode & PE \\ [0.5ex]
\hline
\multirow{4}*{Collaborator A} & \multirow{2}*{Train} & Local & -0.16 \\[-0.1ex]
& & HFL & \textbf{-0.07}  \\[-0.1ex]\cline{2-4}
& \multirow{2}*{Test} & Local & -0.18 \\[-0.1ex]
& & HFL & \textbf{-0.09}  \\[-0.1ex]
\hline

\multirow{4}*{Collaborator B} & \multirow{2}*{Train} & Local & 0.22 \\[-0.1ex]
& & HFL & \textbf{0.13} \\[-0.1ex]\cline{2-4}
& \multirow{2}*{Test} & Local & 0.23 \\[-0.1ex]
& & HFL & \textbf{0.16} \\[-0.1ex]
\hline
\end{tabular}
\caption{Performance of HFL}
\label{tab:HFL}
\end{table}

Table \ref{tab:HFL} provides the evaluations of experiments in HFL collaboration. It can be noted that in the split of the train set of collaborator A, by introducing horizontal collaboration through HFL, the PE is improved from -0.16 to -0.07, while in the test set, it is improved from -0.18 to -0.09. For collaborator B, by working with company A, it can be observed that PE is also improved in both the train set and the test set, from 0.22 to 0.13 and from 0.23 to 0.16, respectively. Thus, collaborator A and collaborator B, can both enhance their claim loss models through an increase in the data volume by privacy-enhancing HFL collaborations, compared to the locally trained models by collaborator A or collaborator B individually.

\begin{table}[!ht]
\centering
\begin{tabular}{c c c rrrr} \hline
Collaborator & Split & Mode & PE \\ [0.5ex]
\hline
\multirow{4}*{Company A} & \multirow{2}*{Train} & Local & -0.16 \\[-0.1ex]
& & VFL & \textbf{0.07}  \\[-0.1ex]\cline{2-4}
& \multirow{2}*{Test} & Local & -0.18 \\[-0.1ex]
& & VFL & \textbf{0.04}  \\[-0.1ex]
\hline
\end{tabular}
\caption{Performance of VFL}
\label{tab:VFL}
\end{table}

Similarly, table \ref{tab:VFL} illustrates the performance of VFL collaboration in our empirical experiments. By performing VFL, collaborator A can utilize the information that it cannot access in the local training mode, so the PE is improved from -0.16 to -0.07, and -0.18 to -0.04, respectively, for the train set and test set. 

By leveraging both HFL and VFL, we have achieved a significant improvement in our performance efficiency (PE) metric. This improvement is directly linked to the business value of our collaborator insurance company, as it has allowed us to enhance the claim loss model. The InsurTech company has also gained valuable insight through our FL collaboration, which has enabled them to provide tailored services to its customers.

\section{Conclusion}\label{sec:future}

In this report, we introduce FL model training to unlock the potential of siloed insurance data without exposing the raw data itself. To date, FL appears to be a largely unexplored technique for insurance industry ML collaborations. Although most insurance companies collect a substantial amount of data, privacy, ethical concerns, sovereignty, and the cost of moving data often result in the data being stored in silos where their use is limited. The use of this data for collaborative training across multiple entities, while protecting the raw data itself from exposure, would be a great benefit to the industry.

Our results in Subsection \ref{subsec:res} demonstrate the use of FL for improved claim loss modeling, by learning a claim loss model from data held at multiple insurance industry collaborators while mitigating exposure of each collaborator's raw data to the others. Our empirical results show a benefit to insurance companies through the use of FL, with more accurate claim loss forecasting. Due to the rarity of claim events among policyholders and the lack of informative rating factors, insurance operations suffer from insufficient data to identify potential risks. Because accurate claim loss modeling can have a significant impact on aspects of businesses, such as underwriting, risk management, solvency, and regulatory compliance, finding solutions to this data shortage problem can be of critical importance.

Although we demonstrate the use of FL for improved loss modeling, there may be other applications of FL in the insurance industry. In addition to fraud detection, other use cases may exist, given the need for business intelligence that arises from the complex structure of the industry. 
\clearpage
\singlespacing
\bibliographystyle{apalike}
\bibliography{ref.bib}

\end{document}